\pdfoutput=1

\documentclass[final]{cvpr}
\usepackage{float}
\usepackage{times}
\usepackage{epsfig}
\usepackage{graphicx}
\usepackage{amsmath}
\usepackage{amssymb}
\usepackage{comment}
\usepackage{bbm}
\usepackage{caption}
\usepackage{xcolor}

\usepackage[pagebackref=true,breaklinks=true,colorlinks,bookmarks=false]{hyperref}

\def\cvprPaperID{2466} 
\def\confYear{CVPR 2021}

\begin{document}

\title{Refer-it-in-RGBD: A Bottom-up Approach for\\ 3D Visual Grounding in RGBD Images}


\author{Haolin Liu$^{1,2}$ \quad\quad Anran Lin$^{1}$ \quad\quad  Xiaoguang Han$^{1,2,}$\thanks{\textbf{Corresponding Email}: hanxiaoguang@cuhk.edu.cn}\quad\quad Lei Yang$^4$ \quad\quad \\Yizhou Yu$^{3,4}$ \quad\quad Shuguang Cui$^{1,2}$\\
$^1$SRIBD, CUHK-Shenzhen\thanks{Shenzhen Research Institute of Big Data} \quad\quad $^2$FNii, CUHK-Shenzhen\thanks{The Future Network of Intelligence Institute, The Chinese University of Hong Kong, Shenzhen} \\$^3$Deepwise AI Lab \quad\quad$^4$The University of Hong Kong
}

\definecolor{Purple}{cmyk}{0.45,0.86,0,0}
\definecolor{Blue}{cmyk}{0.5,0.5,0,0}
\newcommand{\YL}[1]{\textcolor{Purple}{YL: #1}}
\newcommand{\Haolin}[1]{\textcolor{Blue}{Haolin: #1}}
\maketitle

\begin{abstract}

Grounding referring expressions in RGBD image has been an emerging field. We present a novel task of 3D visual grounding in single-view RGBD image where the referred objects are often only partially scanned due to occlusion.
In contrast to previous works that directly generate object proposals for grounding in the 3D scenes, we propose a bottom-up approach to gradually aggregate context-aware information, effectively addressing the challenge posed by the partial geometry.
Our approach first fuses the language and the visual features at the bottom level to generate a heatmap that coarsely localizes the relevant regions in the RGBD image. Then our approach conducts an adaptive feature learning based on the heatmap and performs the object-level matching with another visio-linguistic fusion to finally ground the referred object.
We evaluate the proposed method by comparing to the state-of-the-art methods on both the RGBD images extracted from the ScanRefer dataset and our newly collected SUNRefer dataset. Experiments show that our method outperforms the previous methods by a large margin (by 11.2$\%$ and 15.6$\%$ Acc@0.5) on both datasets.

\end{abstract}
\vspace{-0.5cm}


\section{Introduction}

\begin{figure}[ht]
\centering
\includegraphics[width=.99\linewidth]{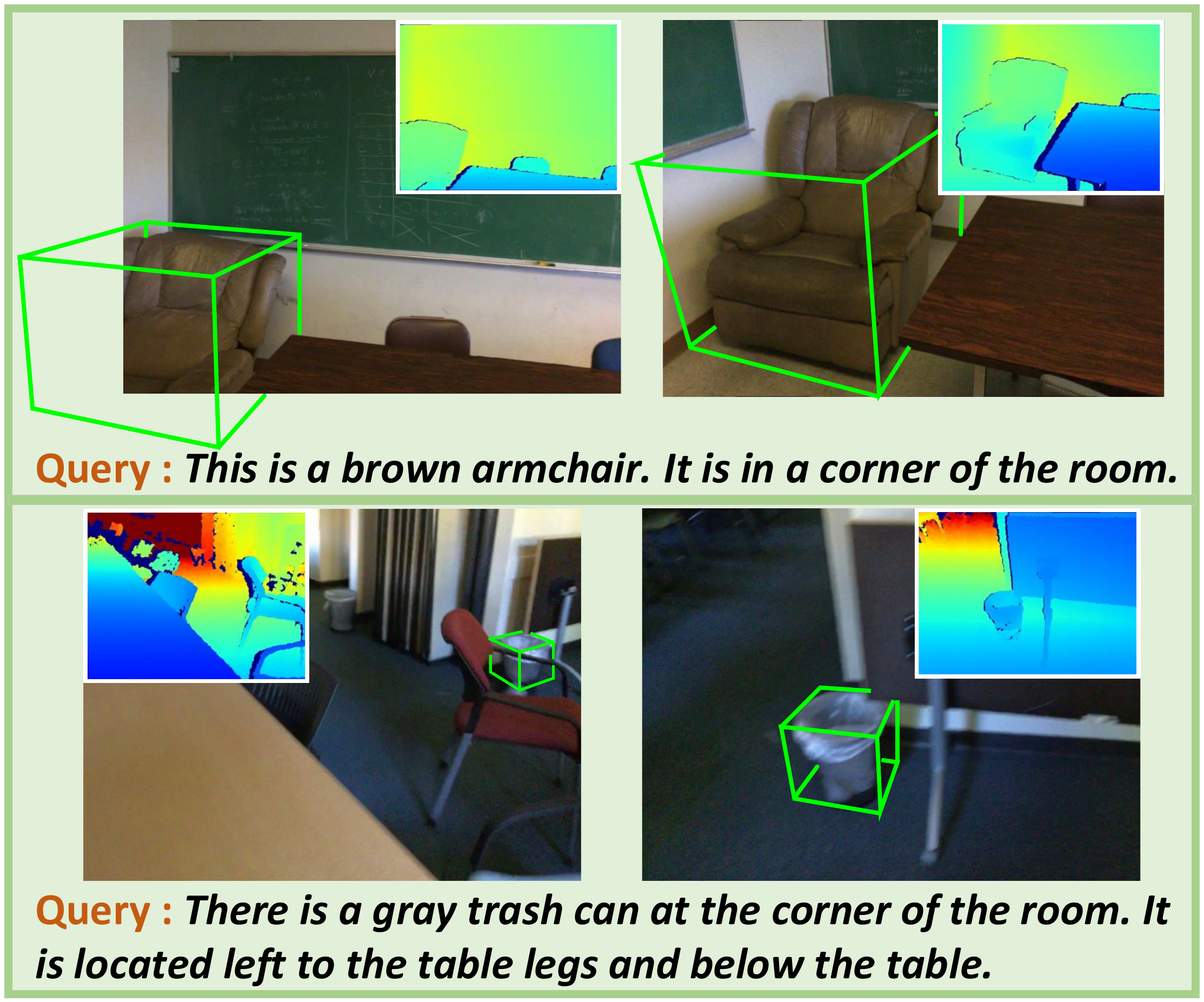}
\caption{\label{fig:teaser}We present a novel task of 3D visual grounding in single-view RGBD images given a referring expression, and propose a bottom-up neural approach to address it. 
Our goal is to estimate the bounding box that encloses the full shape of the referred object even this object is only partially observed (top-left).
Predicted bounding boxes of the referred objects are in green.}
\end{figure}


Localizing objects described by referring expressions in vision signals, also known as \textit{visual grounding}, has long been a major motive for robotics and embodied vision. So far, we have seen growing efforts devoted to visual grounding in images~\cite{krishnavisualgenome,yang2019dynamic,hu2016natural,yu2016modeling,nagaraja2016modeling,rohrbach2016grounding,wang2016learning,chen2017query,yu2017joint,dogan2019neural,zhang2018grounding,deng2018visual,Datta_2019_ICCV,fang2019modularized,liu2019improving,zhuang2018parallel,liao2020real,yang2019cross,yang2020improving,yang2019fast,luo2020multi} and videos \cite{zhou2018weakly,zhou2019grounded,zhang2020object,yang2020weakly,sadhu2020video,shi2019not,zhang2020does}.
Suppose that a robot is going to take `the spoon on the table in the kitchen' following your command ~\cite{janson2018deterministic,misra2015environment}; this would require a more accurate localization result, preferable the 3D coordinate of the referred object rather than a 2D bounding box.
Recent works~\cite{chen2020scanrefer,achlioptas2020referit3d} extend the visual grounding task to 3D scenes~\cite{dai2017scannet} and localize the object referred by a natural language expression. 
While promising results are produced, these methods can only perform 3D visual grounding in complete scenes that are reconstructed and/or segmented~\cite{chen2020scanrefer,achlioptas2020referit3d} in advance. Thus, they are not readily applicable to single-view RGBD images with partial observation, RGBD streaming data, or any dynamically changing environments.


To this end, we propose a novel task for 3D visual grounding: Given a single-view RGBD image of a scene, we aim at estimating the 3D bounding box of the full target object described by a given referring expression. While this novel task opens up many promising possibilities, it also poses a major challenge due to the nature of single-view RGBD images that they contain only incomplete information about the scene and often partial observation of the referred object. Compared to 2D visual grounding on image and 3D visual grounding on complete scene where geometry information is complete in 2D and 3D space, various occlusion cases in our task require a holistic understanding of the object geometry to infer the 3D bounding boxes enclosing the full target object.

Image-based grounding methods may be applied to RGBD images but require an extra effort to lift the produced 2D bounding boxes to 3D. Chen et al.~\cite{chen2020scanrefer} proposed a one-stage search and match strategy for 3D grounding. However, it fails to handle single-view RGBD images where the referred objects are partially observed. 
The reason is two-fold: 
First, it is inadequate to directly match between features of these object proposals and the referring expression to achieve reliable grounding, as each proposal contains only incomplete information due to the partial observation. Moreover, this is worsened by the fact that only content-free object proposals are generated by a detection network that searches the scene globally, thus failing to accumulate useful information about the referred object for grounding.

With these observations, we propose a novel, bottom-up matching approach for fine-grained grounding of the referred objects in given single-view RGBD images. 
To this end, our approach first matches the query expression to the input RGBD image and generates a content-aware heatmap on the voxel domain converted from the RGBD image. This bottom-level matching amounts to coarse localization of regions, which are relevant to the referred object.
Then, based on the content-aware coarse localization, an adaptive search-and-match strategy is employed. This enables our network to conduct fine-grained search in the relevant regions and generate visio-linguistic features by fusing the query with more informative features from the visual modality. These fused features are used to generate and refine the 3D object proposals to the final bounding box enclosing the target object in the given referring expression. 
Compared to previous works, our bottom-up approach exploits the language features at different levels, and thus enables our network to be content-aware during searching and matching stages. The adaptive search guided by the content-aware heatmap also ensures the feature learning to be concentrated in the relevant regions, mitigating the challenge posed by the partial geometry.

Mauceri et al.~\cite{mauceri2019sun} present the SUN-Spot dataset that provides spatial referring expressions to raw single-view RGBD images in the SUNRGBD dataset\cite{song2015sun}. However, the amount of language annotations as well as their linguistic variations (only spatial references) are inadequate. Alternatively, one can extract the RGBD frames from 3D scene datasets~\cite{Matterport3D,qi2020reverie,dai2017scannet,chen2020scanrefer,achlioptas2020referit3d} that also provide rich object-centric language descriptions. Yet, referring expressions provided in these scene datasets are constructed based on the scene context; they may contain other supporting objects that exist in the scene but are not observed in a particular frame. 
This artifact between annotations and the extracted frames from the 3D scenes motivates us to contribute a large-scale annotation dataset, SUNRefer, to facilitate future studies on visual grounding in single-view RGBD images. Built on the SUNRGBD dataset, our dataset contains 7,699 RGBD images with a total of 38,495 annotations of referring expressions on 7,699 objects.

We evaluated our proposed 3D visual grounding method on both SUNRefer, our newly constructed dataset, and the ScanRefer dataset using the extracted RGBD frames. We show that our approach outperforms the state-of-the-art methods by a significant margin and validate our desgin choices via extensive ablation study. Our method can also be applied to 3D visual grounding in streaming RGBD images at a processing rate of 10 frame-per-second.

Our key contributions are summarized as follows:

1) We present a novel and challenging task -- 3D visual grounding in single-view RGBD images with possible incomplete information or partial occlusion 

2) We propose a content-aware, bottom-up approach that significantly outperforms the state-of-the-art methods on both our newly collected dataset and the ScanRefer dataset.

3) We contribute a large-scale dataset of referring phrases and the corresponding ground-truth bounding boxes for a large amount of publicly available RGBD images.

\section{Related Work}
\begin{figure*}[ht]
    \centering
    \includegraphics[width=.99\linewidth]{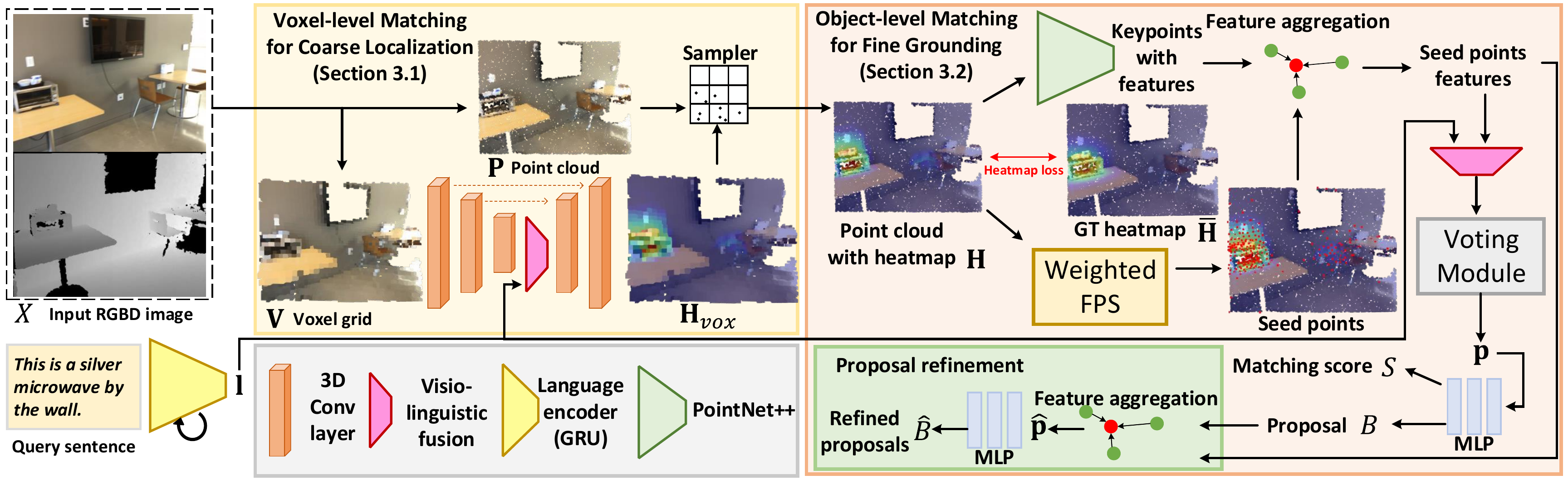}
    \caption{Overview of the proposed method. Our bottom-up approach consists of two main modules. Given an RGBD image and a referring sentence, we first match the input pair at the voxel level to coarsely localize the relevant regions. This is followed by an adaptive feature learning enabled by weighted FPS based on the coarse localization result. An object-level matching between the language feature and the context-aware visual features is performed to generate object proposals for fine grounding. Finally, the referring sentence is grounded in the object proposal with the highest score. 
    }
    \label{pipeline}
\end{figure*}

Given a referring expression along with an input image, 2D visual grounding tasks aim to estimate a 2D bounding box or instance segmentation mask that localizes in the image the object most relevant to the referring expression. 
Previous works \cite{hu2016natural,yu2016modeling,nagaraja2016modeling,rohrbach2016grounding,wang2016learning,chen2017query,yu2017joint,dogan2019neural,zhang2018grounding,deng2018visual,Datta_2019_ICCV,fang2019modularized,liu2019improving,zhuang2018parallel,yang2019dynamic} usually adopt a two-stage approach. Firstly, object proposals or segmentation masks are produced by pre-trained models. Then, these methods rank the matching scores between the referring expression and the proposals to retrieve the best-matched object proposal as the grounding result. However, performance of the two-stage approach is bounded by the pre-trained models chosen. 
one-stage methods are also proposed~\cite{liao2020real,yang2019cross,yang2020improving,yang2019fast,luo2020multi} which produce both object proposals and matching scores. 
In addition, similar approaches can be applied to object visual grounding in streaming video frames~\cite{zhang2020object,yang2020weakly,sadhu2020video,shi2019not,zhang2020does,zhou2018weakly,zhou2019grounded} to ground objects or referring expressions in videos.

However, visual grounding in 2D images is limited since it is only able to localize the referred objects with 2D bounding boxes in the image domain. It is still unavailable to get the precise 3D locations of the objects which are desirable for more advanced tasks, such as embodied AI\cite{qi2020reverie,majumdar2020improving,anderson2018vision,wijmans2019embodied,das2018embodied}.

Grounding texts in the indoor scenes has been an important research field. An earlier work~\cite{kong2014you} explored the text-to-image coreference in RGBD scenes, which takes sentences describing the whole scene as input and matches each noun word with an object in the scene. This differs from ours task where input sentences are referring to a particular object in the scene.
Qi et al.~\cite{qi2020reverie} proposed a language-guided navigation task in the indoor scenes where an agent needs to follow the language instructions to find a remote target. In this study, we focus on the task of 3D visual grounding, i.e., localizing the 3D bounding box of the referred object in the single-view RGBD images. 
A close related work is ScanRefer~\cite{chen2020scanrefer} where the authors propose a 3D visual grounding task to find the 3D bounding boxes of referred objects in a complete RGBD scene which requires reconstruction from raw scans beforehand. ReferIt3D~\cite{achlioptas2020referit3d} is another recent work on 3D visual grounding in scenes. This work assumes the availability of segmentation masks of object instances for both training and testing.
We are different to both previous works in that we focus on grounding in single-view RGBD images where partial occlusions or incomplete data are our major challenges.

We also differ from ScanRefer in terms of methodology. ScanRefer employs a one-stage matching between the object proposals generated by \cite{qi2019deep} and the language features to obtain the grounding results. In contrast, we propose to conduct 3D visual grounding using a bottom-up approach. We first conduct a visual-language matching on the voxel domain to coarsely localize the relevant regions. Then, we accumulate features from these relevant regions for the following fine-grained matching on the object level.

\section{Method}

\textbf{Overview}
We present a novel task of grounding a referring expression of a 3D object in a given RGBD image. The expected output of this task is the 3D bounding box of the full object referred in the language expression. This is not a trivial task as the input RGBD image may contain only part of the referred object. 
We propose a novel neural solution to tackle this task in a bottom-up approach. Our framework consists of two main modules as shown in Figure~\ref{pipeline}. We first produce a confidence heatmap from voxel-level matching between voxel and the query sentence, to coarsely localizes the relevant regions to the referred object in the image. Next, an adaptive search strategy is proposed to conduct fine-grained, content-aware search for the referred object based on the heatmap. Finally, we fuse the features accumulated from adaptive search and the language feature to conduct object-level matching, yielding matching scores and axis-aligned bounding box.

\subsection{Voxel-level Matching for Coarse Localization}

Given a referring expression and an RGBD image, this module aims to learn from the two input modalities and produces a heatmap using the global context to localize relevant regions to the referred object in the input image. 

\textbf{Heatmap generation. }
The input RGBD image $\mathbf{X}$ is converted to a voxel grid $\mathbf{V}$ with $0.05 m$ cell size.
We adopt a U-Net model~\cite{rohrbach2016grounding}, an encoder-decoder structure with skip links constructed by 3D sparse convolutions~\cite{choy20194d}, to regress a voxel-wise heatmap $\mathbf{H}_{vox}$ indicating the relevancy of the referred object in the voxel grid.
For further grounding in the point cloud data, we project the produced heatmap $\mathbf{H}_{vox}$ to the point cloud $\mathbf{P}$, and eventually yield a heatmap $\mathbf{H}$ on the point cloud domain $\mathbf{P}$.

The voxel grid is chosen to produce the heatmap due to two considerations. First, they can maintain the spatial awareness of the objects compared to the 2D images. Second, they are insensitive to superficial details compared to the point clouds. This design choice is further validated in our ablation study (see Table~\ref{table_abalation} and Figure~\ref{diff_modality}).

\textbf{Fusing language features for coarse localization. }
To fuse the language input for the grounding purpose, We first pass the given referring expression to a language encoder and obtain language feature $\mathbf{l}$. In particular, We employ GloVe~\cite{pennington2014glove} to obtain the vector embedding for each word, and use Gated Recurrent Units (GRU) to extract feature $\mathbf{l}$ from the sequence of word embedding vectors.

We then introduce language feature $\mathbf{l}$ to modulate the heatmap generation process for content-aware coarse localization. Specifically, we concatenate the output feature maps of the voxel encoder with $\mathbf{l}$, and feed the concatenated features to a visio-linguistic encoder to obtain the fused features. It is followed by the voxel decoder that generates the content-aware heatmap $\mathbf{H}_{vox}$ that coarsely localizes the regions relevant to the referring expression.



\subsection{Object-level Matching for Fine Grounding}

We leverage the generated heatmap to perform fine-grained grounding by a high-level object and language matching. To this end, we first propose a content-aware sampling to generate seed points that concentrated in the regions highlighted by the heatmap.
We then conduct a visio-linguistic fusion between the seeds and language feature, and employ a voting mechanism proposed in~\cite{qi2019deep} for proposals generation and object-language matching.

\textbf{Adaptive feature learning. }
Based on the heatmap, we conduct an adaptive feature learning using PointNet++\cite{qi2017pointnet++} to aggregate information only at the relevant regions. 

To this end, we employ a weighted farthest point sampling (weighted FPS) with a modified distance metric to obtain adaptive samples guided by the heatmap. In particular, we use the heatmap value at the query point $\mathbf{q}$ as the scaling factor, and modify the Euclidean distance metric used in farthest point sampling (FPS) as follow:
\begin{equation}
    \hat{d}(\mathbf{q}, \mathbf{c})=h(\mathbf{q}) d(\mathbf{q}, \mathbf{c})
\end{equation}
where $d(\mathbf{q}, \mathbf{c})$ is the Euclidean distance between points $\mathbf{q}$ and $\mathbf{c}$. $h(\mathbf{q})$ is the heat value of point $\mathbf{q}$. 
This weighted FPS can ensure that a point with a higher heat value will have a higher chance to be chosen, and thus result in seed points densely distributed in relevant regions identified by the heatmap. The weighted FPS also maintains the uniformity of the sampled points in the relevant region (as shown in Figure ~\ref{diff_sampling}).

The adaptive sampled seed points are denoted as  $\{\mathbf{s}_i\}_{i=1}^{M}$. 
    To obtain the features associated with the seed points, PointNet++\cite{qi2017pointnet++} is first employed to extract features (residing in a set of key-points) from point cloud $\mathbf{P}$. 
We then use the Feature Propagation module as in \cite{qi2017pointnet++} to aggregate features from the nearby key-points to our seed points. 

\textbf{Visual-language fusion for proposal generation. }
Next, we fuse the language features and visual features sampled in a content-aware manner for the object-level grounding. 
Specifically, we concatenate seed point features $\mathbf{f}_i$ with the language feature $\mathbf{l}$, and process them using another visio-linguistic encoder for fusion.
\begin{equation}
    \mathbf{\hat{f}}_i = E_{VL}(cat(\mathbf{f}_i, \mathbf{l})).
\end{equation}
We denote the seed points with associated fused features as $\{\mathbf{\hat{s}}_i = [\mathbf{s}_i \in \mathbb{R}^3,~ \mathbf{\hat{f}}_i \in \mathbb{R}^C]\}_{i=1}^{M}$.

We adopt the voting module proposed in \cite{qi2019deep} to generate votes $\{\mathbf{v}_i\}_{i=1}^{M}$ pointing at objects' center from the obtained seed points $\{\mathbf{\hat{s}}_i\}_{i=1}^{M}$. 
Then, the votes are clustered in the 3D space to produce object proposal features $\{\mathbf{p}_k \in \mathbb{R}^{3+C}\}_{k=1}^{K}$, where the first three channels represent the cluster center of the proposal. 
Finally, we use a shared MLP-based regressor to estimate, for each of the $K$ proposal features, the proposal box $B_k = (x,y,z,w,l,h)$ and its score $S_k$ to match the referred object.
Among $K$ candidate proposals, we choose the highest scored proposal as our predicted bounding box to localize the referred object. 

\textbf{Proposal refinement: }
As the proposal features are aggregated from each cluster formed by the votes, they contain only local information from that cluster. Thus, we provide an optional module to refine the proposal features $\{\mathbf{p}_k\}_{k=1}^{K}$. We perform another feature aggregation among $\{\mathbf{p}_k\}_{k=1}^{K}$ and $\{\mathbf{s}_i\}_{i=1}^{M}$ to produce a set of refined proposal features $\{\mathbf{\hat{p}}_k\}_{k=1}^{K}$. Then we feed $\{\mathbf{\hat{p}}_k\}_{k=1}^{K}$ to the regressor to produce final proposal boxes $\{\hat{B}_{k}\}_{k=1}^K$.

\begin{table*}[ht]\footnotesize
	\begin{center}
		\caption{Statistics of SUNRefer dataset comparing some public 3d referring expression datasets.}
		\label{dataset}
		\resizebox{.8\linewidth}{!}{%
			\begin{tabular}{c|c|c|c|c}
				dataset & $\#$ of annotations &$\#$ of objects& Annotated on & averaged description length\\
				\hline
				ScanRefer~\cite{chen2020scanrefer} & 51,583 & 11,046 & 704 scenes & 20.3\\
				Nr3D~\cite{achlioptas2020referit3d} &41,503  & 5,879 & 642 scenes & 11.4\\
				REVERIE~\cite{qi2020reverie} & 21,702 & 4,140 & 90 scenes & 18.0\\
				SUN-Spot~\cite{mauceri2019sun} & 7,990 & 3,245 & 1,948 RGBD images  & 14.1\\
				SUNRefer & 38,495 & 7,699 & 7,699 RGBD images & 16.3 \\
				\hline
			\end{tabular}
		}
	\end{center}
\end{table*}

\subsection{Loss Function}

We train our network with the following loss function,
\begin{equation}
\label{total loss}
\begin{aligned}
    \mathcal{L}_{total}&=\lambda_1\mathcal{L}_{heat}+\lambda_2\mathcal{L}_{vote}+\lambda_3\mathcal{L}_{match}+\lambda_4\mathcal{L}_{resp}
\end{aligned}
\end{equation}

Heatmap loss $\mathcal{L}_{heat}$ encourages the generation of the distinctive heatmap which is able to determine regions' relevancy to the sentence query. We firstly construct a ground-truth (GT) point cloud heatmap $\bar{\mathbf{H}}$ according to the target object's location. We calculate it using a Gaussian kernel so that further distance from the target object's center will have diminished value. $\mathcal{L}_{heat}$ is obtained by the mean square error (MSE) between $\bar{\mathbf{H}}$ and generated $\mathbf{H}$ as follow:
\begin{equation}
    \mathcal{L}_{heat}=\left\|\bar{\mathbf{H}}-\mathbf{H}\right\|_2
\end{equation}

Vote loss $\mathcal{L}_{vote}$is defined following~\cite{qi2019deep}:
\begin{equation}
    \mathcal{L}_{vote}=
    \frac{1}{M_{pos}}\sum_{i}\left\|v_i-\bar{v}_i\right\|_2\mathbbm{1}[s_i~\text{on objects}],
\end{equation}
where $M_{pos}$ is the total number of seed points on objects, $v_i$ and $\bar{v}_i$ are the predicted vote coordinates and GT centers of the object that the seed point $s_i$ resides on, if applicable. Thus, the vote loss enforces the network to produce a correct vote, when a seed point on any object is given.

Matching loss $\mathcal{L}_{match}$ is designed to ensure matching score $S_k$ to approximate the Intersection of Union (IoU) between the proposal $B_k$ and the GT bounding box of the target object $\bar{B}$. This way, it can return high responses to correct proposals while suppressing any mismatched proposals. Thus, the matching loss is computed using MSE as follows:
\begin{equation}
    \mathcal{L}_{match}=\frac{1}{K}\sum_{i=1}^K\left\|IoU_k-S_k\right\|_2
\end{equation}
where $K$ is the number of proposals and $IoU_k$ denotes the IoU between $B_k$ and $\bar{B}$.

Response loss $\mathcal{L}_{resp}$ aims to pull the proposal $B_i$ that has the maximum IoU with GT bounding box closer to it. Given target bounding box $\bar{B}$, The loss is written as follow:
\begin{equation}
    \mathcal{L}_{resp}=\left\|B_i-\bar{B}\right\|_2, \quad i=\arg\max_k (IoU_k).
\end{equation}

When proposal refinement is employed, $\{B_k\}_{k=1}^K$ will be supervised by GT bounding box $\bar{B}^{partial}$ enclosing the partial object in the input and the refined proposals $\{\hat{B}_k\}_{k=1}^K$ will be supervised by GT bounding box $\bar{B}^{intact}$ enclosing the full object. Otherwise, Proposals $\{B_k\}_{k=1}^K$ will be directly supervised by $\bar{B}^{intact}$.
\section{SUNRefer Dataset}

Due to the lack of a large-scale dataset dedicated to visual grounding in single-view RGBD images, we contribute a large-scale referring expression dataset, named SUNRefer. 
Our dataset contains 7,699 RGBD images from SUNRGBD~\cite{song2015sun} and annotation of 3d orientated bounding boxes enclosing the full objects. We hire annotation workers to annotate each target object with referring expressions.  

Given an RGBD image with the bounding box enclosing an object, we ask the annotators to describe this object using its own attributes (e.g. color, shape, and material) and/or spatial relationship in the surrounding environment with multiple sentences. A description is expected to distinguish the target object from its neighboring objects and those with similar appearances but different locations. 

\begin{figure}[h]
    \centering
    \includegraphics[width=.99\linewidth]{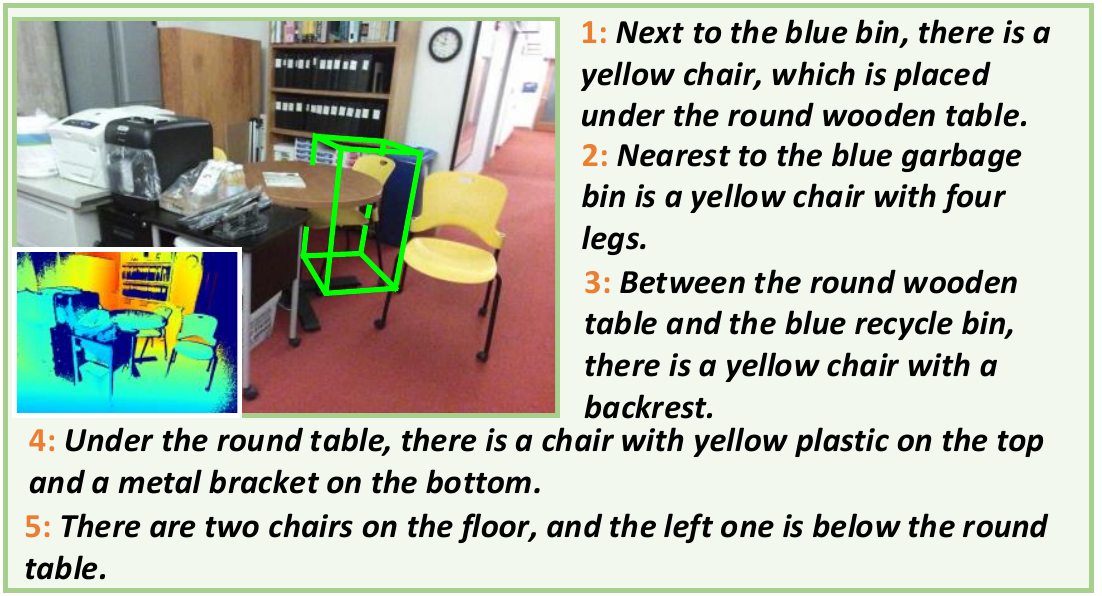}
    \caption{An example of the SUNRefer dataset with five different language descriptions referring to the chair enclosed by the green bounding box.}
    \label{fig:dataset_example}
    \vspace{-4mm}
\end{figure}

For each target object, we collect five descriptions from different annotators in order to ensure linguistic diversity. Verification was conducted to obtain high-quality annotations. 
An example of SUNRefer dataset is shown in Figure~\ref{fig:dataset_example}. Statistics of the SUNRefer dataset and comparison with some public datasets are shown in Table~\ref{dataset}. More details can be found in our supplementary.
\section{Experiments}
\begin{table*}[ht]\footnotesize
	\begin{center}
		\caption{Quantitative comparison between our methods and the state-of-the-art methods. Our method outperforms the comparing methods by a large margin on both SUNRefer and ScanRefer datasets under the single-view RGBD setting.}
		\label{table_stoa}
		\resizebox{.7\linewidth}{!}{%
			\begin{tabular}{c|ccc|ccc}
			    Dataset & \multicolumn{3}{c|}{SUNRefer}& \multicolumn{3}{c}{ScanRefer (single-RGBD)}\\
				\hline
				Methods & Acc@0.5 & Acc@0.25& R@5& Acc@0.5& Acc@0.25 & R@5\\
				\hline\hline
				ReSC\cite{yang2020improving}  &7.4  &23.6  &9.4 &12.4 &34.3 &17.1\\
				One Stage\cite{yang2019fast}  &2.6 &12.8 &7.9 &5.8 &26.1 &21.7\\
				ScanRefer\cite{chen2020scanrefer}  &20.3 &45.0 &23.6 &20.3 &48.2 &33.8 \\
				Ours &\textbf{35.9} &\textbf{49.6} &\textbf{52.9} & \textbf{31.5}&\textbf{56.5} &\textbf{49.3}\\
				\hline
			\end{tabular}
		}
	\end{center}
\end{table*}

\begin{table}[h]\footnotesize
	\begin{center}
		\caption{Quantitative comparison against the state-of-the-art methods on the ScanRefer dataset. Our method achieves comparable, if not better, results comparing to two variants of ScanRefer even under the whole-scene setting.}
		\label{table_whole}
		\resizebox{.9\linewidth}{!}{%
			\begin{tabular}{c|ccc}
			    Dataset & \multicolumn{3}{c}{ScanRefer (whole scene)}\\
				\hline
				Methods & Acc@0.5 & Acc@0.25 & R@5\\
				\hline\hline
				ScanRefer-full  &27.1 &\textbf{41.0} &43.5\\
				ScanRefer-xyz+rgb  &23.1 &35.9 &39.7\\
				FPS baseline-xyz+rgb &24.2& 35.1& 44.7\\
				Ours-xyz+rgb &\textbf{29.0}&40.2 &\textbf{47.1}\\
				\hline
			\end{tabular}
		}
	\end{center}
\end{table}
\begin{table*}[ht]\footnotesize
	\begin{center}
		\caption{Quantitative results of ablation study. We compare our full model using the voxel-based heatmap and the weighted FPS with several alternative configurations. The comparisons validate our design choices.}
		\label{table_abalation}
		\resizebox{.8\linewidth}{!}{%
			\begin{tabular}{c|ccc|ccc}
			    Dataset & \multicolumn{3}{c|}{SUNRefer}& \multicolumn{3}{c}{ScanRefer (single-RGBD)}\\
				\hline
				Methods & Acc@0.5 & Acc@0.25& R@5& Acc@0.5& Acc@0.25 & R@5\\
				\hline\hline
				Ours (voxel + weighted FPS) &\textbf{35.9} &\textbf{49.6} &\textbf{52.9}&\textbf{31.5}&\textbf{56.5} &\textbf{49.3}\\
				Voxel-level matching only &12.2&31.0&18.4 & 16.8& 39.8&28.4\\
				FPS baseline w/o voxel-level matching &28.6&45.7&43.5&25.4 &52.7 &39.2\\
				Weigthed random sampling &34.1&48.2&51.9& 29.4&56.4&46.8\\
				\hline
				image modality &33.6&48.3&48.3& 29.2&54.9& 45.1\\
				point cloud modality &34.0&47.8&50.4&29.8&55.2&48.3\\
				\hline
				w/o proposal refinement &35.8 &49.5 &52.9 &30.7&55.6&47.7\\
				\hline
			\end{tabular}
		}
	\end{center}
	\vspace{-0.5cm}
\end{table*}
\begin{figure*}[ht]
    \centering
    \includegraphics[width=.90\linewidth]{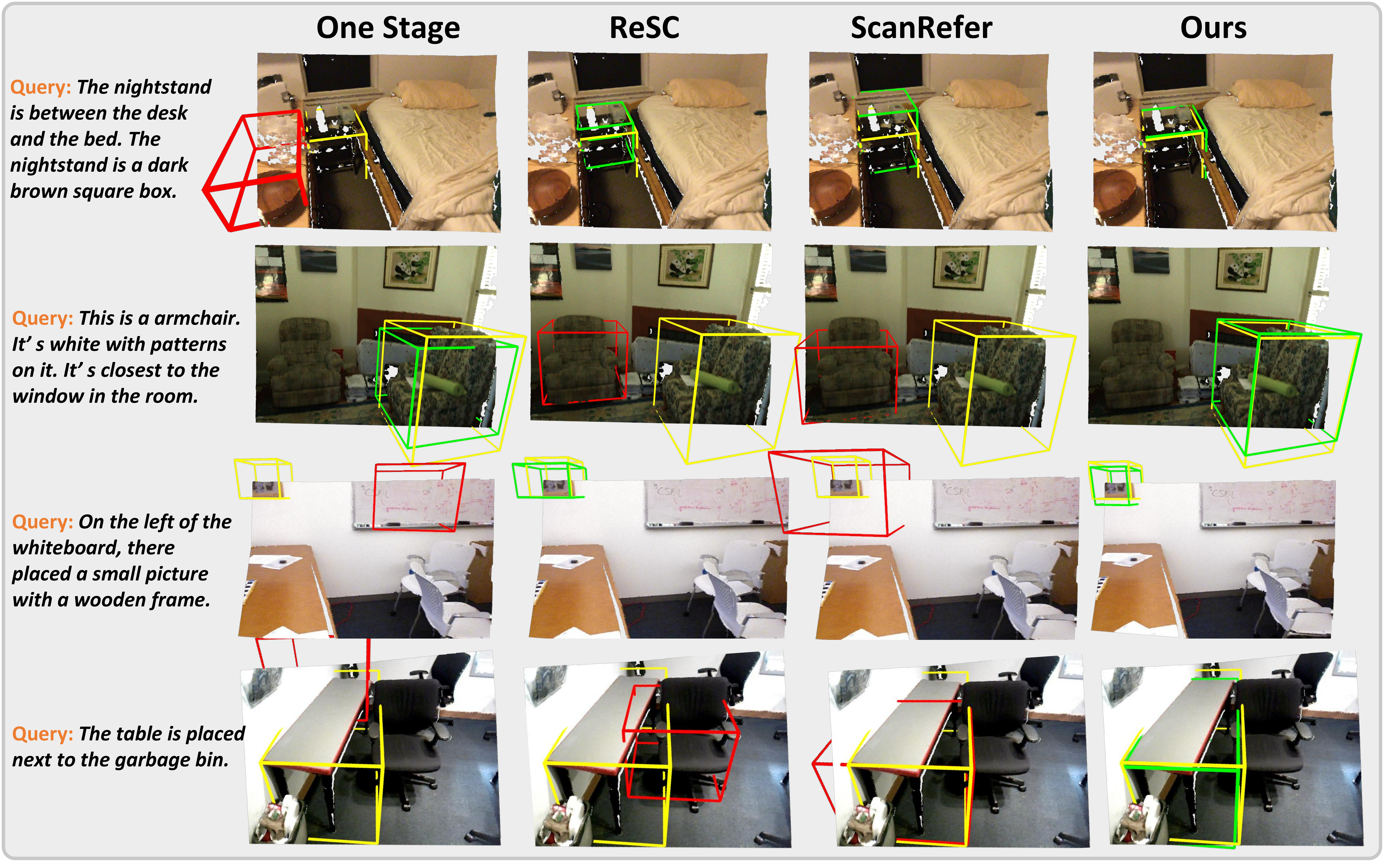}
    \caption{Comparison between our method and the state-of-the-art.
    A bounding box is considered as a successful prediction (green) if it has an IoU larger than 0.5 with the ground-truth box (yellow) of the entire object even if it is partially observed; otherwise, it is considered as a failure one (red). More visual results are presented in the supplementary.}
    \label{fig:compare_stoa}
\end{figure*}

\subsection{Implementation Details}

We first train the language-guided coarse localization module (and the GRU) independently with the heatmap loss for 10 epochs. Based on this pre-trained module, we then train the entire framework with the described loss function (Equation~\ref{total loss}) for another 60 epochs.
The coefficients for the loss terms are $\lambda_1=0.5,\lambda_2=0.5,\lambda_3=2.0,\lambda_4=1.0$. 
The Adam optimizer~\cite{kingma2014adam} is used for optimization. The learning rate is set to $0.0001$ for the coarse localization module, and $0.001$ for the rest of the model parameters. Both learning rates are decayed by $0.2$ at the 50-th epoch. 
Data augmentation is applied to both voxel and point cloud models by perturbing them with small translations and rotations in all three axes. 
The batch size is set to $14$ for training.
The training time is approximately 30 hours on a single NVIDIA RTX-2080-TI GPU.
More implementation details are supplied in the supplemenetary materials

\subsection{Evaluation Metric}

We use the highest scored proposal from our network as the predicted bounding box of the referred object. We follow~\cite{chen2020scanrefer} to use Acc@0.5 and Acc@0.25 as our evaluation metrics.
We also introduce another evaluation metric, R@5, which is widely used in 2D visual grounding to reflect the retrieval accuracy, where the retrieval is considered correct if at least one of the proposals of top-5 scores has an IoU $>0.5$ with the GT box. 
Thus, Acc@$\{0.5,~0.25\}$ focus on how close the highest scored proposal to the GT box, and R@5 evaluates the quality of the proposal candidates.
\subsection{Comparison with SOTA}

\textbf{Comparison in the single-view RGBD setting: }
We compare the proposed method with ScanRefer~\cite{chen2020scanrefer} and the other two methods extended from 2D grounding, i.e., One Stage~\cite{yang2019fast} and ReSC~\cite{yang2020improving}, under the single-view RGBD setting on both ScanRefer and SUNRefer datasets.

As the ScanRefer dataset provides only the complete scene annotations, we extract the raw RGBD images from ScanNet~\cite{dai2017scannet}. For each extracted images, we identify the objects that are visible in the image and use these objects and their corresponding referring sentences as paired input to our network for training and testing. We select a subset of the images in which the objects have different visibility ranging from high to nearly invisible. This ensures our model can cope with target objects with severe incompleteness. We follow the train/val split of the ScanRefer dataset.


For the two methods extended from 2D grounding, we train a VoteNet to generate 3D object proposals and match them with the 2D grounding results to produce the 3D grounding output. See Supplementary for more details.

Quantitative comparison between these methods is shown in Table~\ref{table_stoa}. Our method significantly outperforms the other methods with a large margin on both datasets. This shows that ScanRefer may not be suitable for grounding in single-view RGBD images. Performance of 2D visual grounding methods indicates the inefficiency of obtaining 3D precise localization results from a 2D bounding box. 
Figure~\ref{fig:compare_stoa} depicts some qualitative results. The proposed method has both higher retrieval accuracy and localization accuracy than previous methods.  

\textbf{Comparison in the whole-scene setting: } 
We then discuss the performance of our method in the whole-scene setting, the same setting as in~\cite{chen2020scanrefer}. We compare our method to two ScanRefer variants: 1) \textit{ScanRefer-xyz+rgb} using the spatial coordinates and colors; and 2) \textit{ScanRefer-full} using its full configuration (including pre-trained multi-view image features, point cloud normals, and a language classifier). 
Table~\ref{table_whole} shows our method outperforms the two variants of ScanRefer in Acc@0.5 (29.0 against 27.1) and R@5 (47.1 against 43.5), and attains a comparable performance in Acc@0.25. Our method using xyz+rgb outperforms ScanRefer with the same input and achieves comparable results compared to the full configuration of ScanRefer which uses extra inputs. This shows our bottom-up grounding method is applicable to the 3D scenes as well.

\begin{figure}[h]
    \centering
    \includegraphics[width=.99\linewidth]{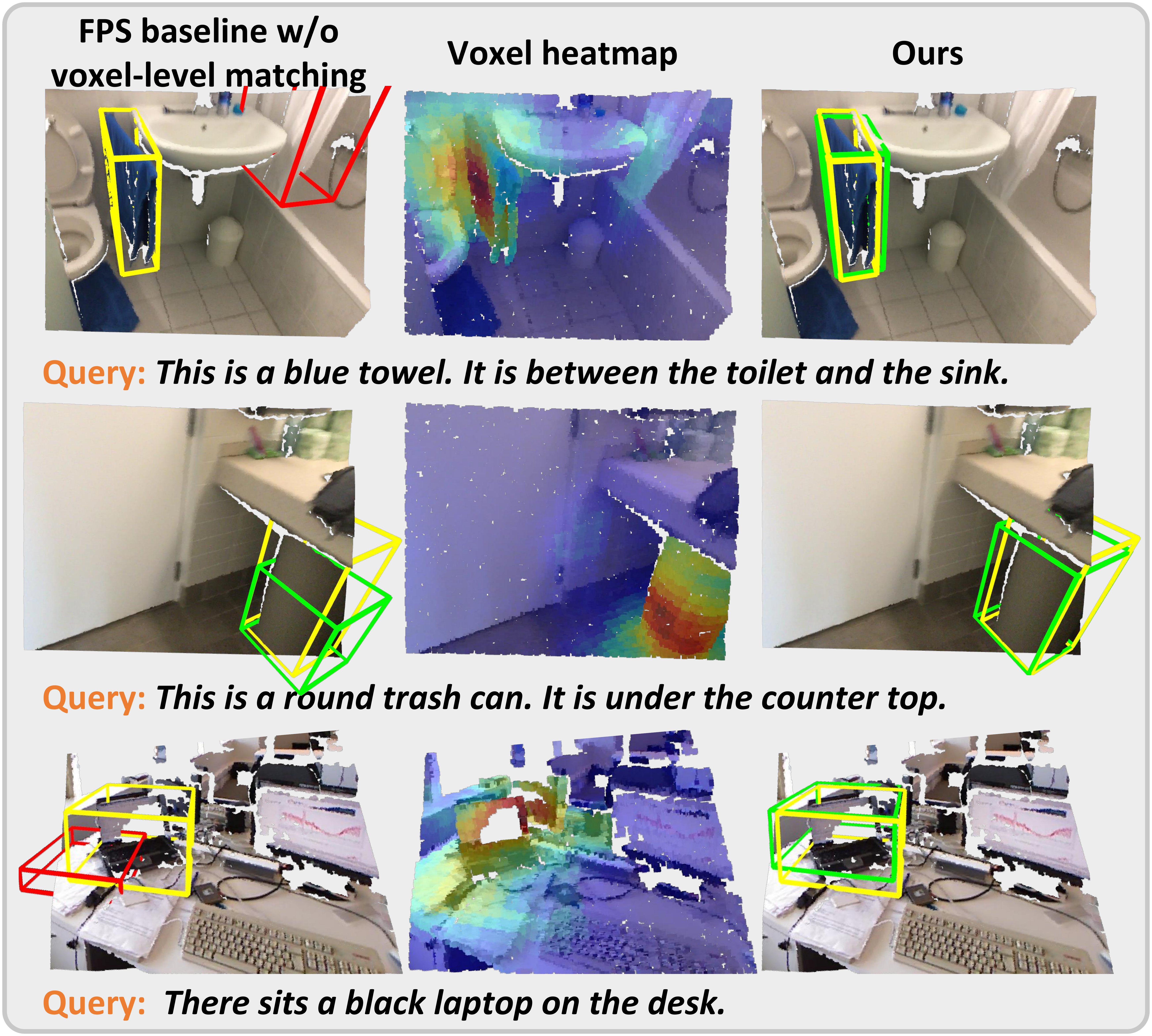}
    \caption{Comparison between the FPS baseline with the proposed bottom-up approach. Successful predictions are in green; failures in red; and ground-truth in yellow.}
    \label{baseline_vox}
\end{figure}

\subsection{Ablation Study}

In order to show the effectiveness of different design choices in our proposed approach, we conducted extensive ablation studies.

\begin{figure}[h]
    \centering
    \includegraphics[width=.90\linewidth]{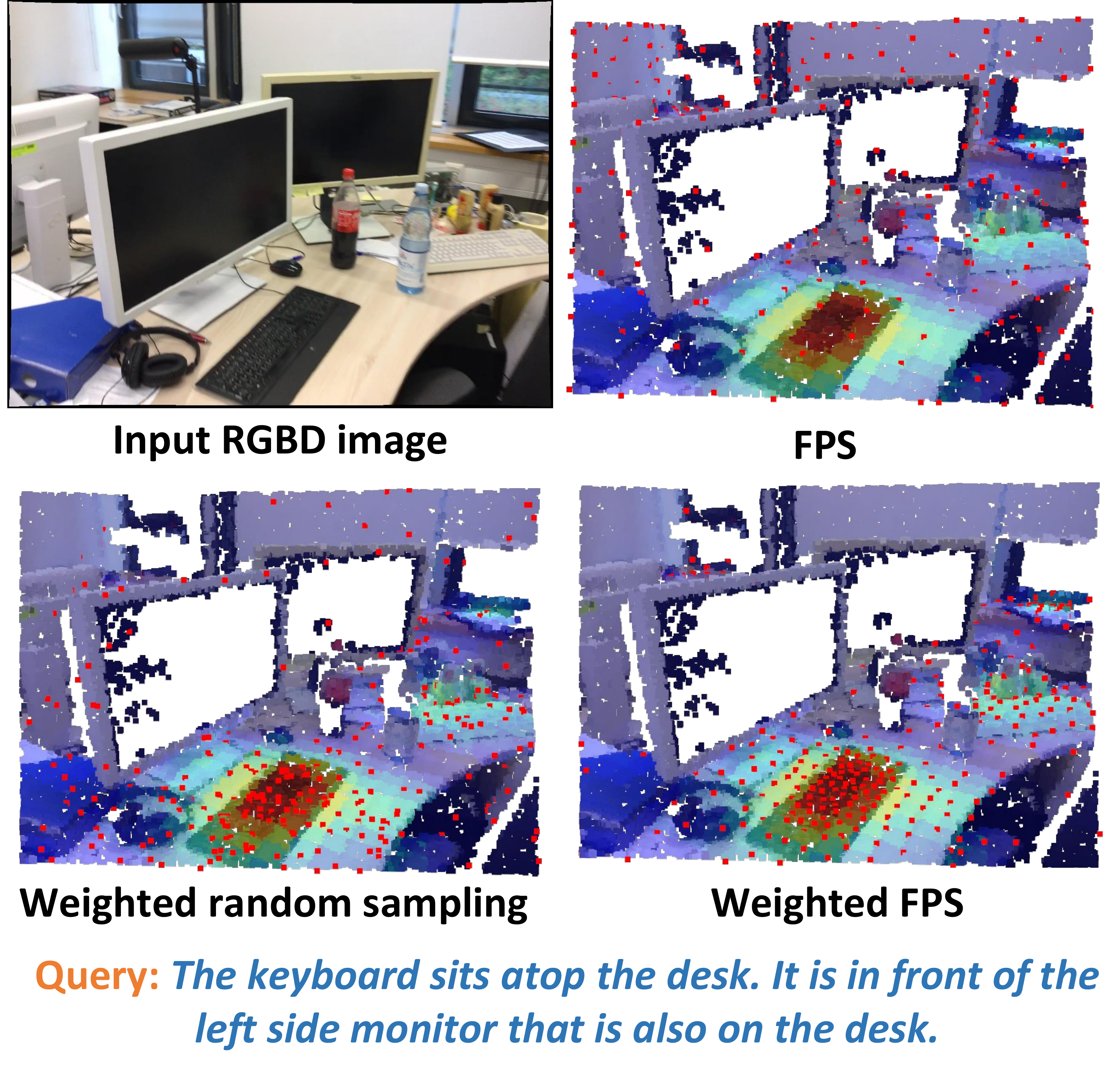}
    \caption{Comparison between different sub-sampling methods. Seed points are visualized in red.}
    \label{diff_sampling}
\end{figure}

\textbf{FPS baseline w/o voxel-level matching:}
We first compare our method to the FPS baseline where no voxel-level matching is performed. Thus, the FPS baseline amounts to use a standard VoteNet~\cite{qi2019deep} with a single visual-language matching at the object level.
Quantitative comparison demonstrates that our method with voxel-level matching is able to achieve far better results; see the first row block of Table~\ref{table_abalation}. 
Some qualitative results are provided in Figure~\ref{baseline_vox}. In the first and third row, our method successfully localizes the correct object, while FPS baseline fails. In the second row, both methods are successful, yet our method obtain more accurate results.

\textbf{Voxel-level matching only:}
We also consider the configuration where object-level matching is removed. We conducted an experiment of using only voxel-level matching for grounding. Specifically, we trained a VoteNet~\cite{qi2019deep} to generate object proposals and ground the text in the proposal with the highest confidence value based on the point cloud heatmap. Results are shown in the first row block in Table~\ref{table_abalation}. Using only voxel-level matching severely deteriorates the performance, further demonstrating the necessity of combining voxel-level and object-level matching.

\textbf{Different sampling strategies:}
We show the advantage of weighted FPS over the weighted random sampling technique which randomly samples the point set using their heatmap value as the sampling probability.
As can be seen from the first row block of Table~\ref{table_abalation}, weighted FPS (Ours) can attain slightly better results compared to weighted random sampling. Both methods outperform the FPS baseline by a large margin.
The performances show that weighted FPS with contextual information facilitates object proposal generation. 
Intermediate results are shown in Figure~\ref{diff_sampling} where weighted FPS can yield well-patterned seed points that are densely concentrated in the relevant regions. The weighted random sampling only scatters the majority of seed points around relevant regions, while the FPS spreads the seed points uniformly in the point cloud with few seed points in the relevant regions. 


\textbf{Different modalities for heatmap generation:}
We validate the design choice of using voxel modality for coarse localization. We compare our model using 3D voxels with its variants using 2D images and 3D point clouds. See supplementary for more details of these variants.

\begin{figure}[h]
    \centering
    \includegraphics[width=0.9\linewidth]{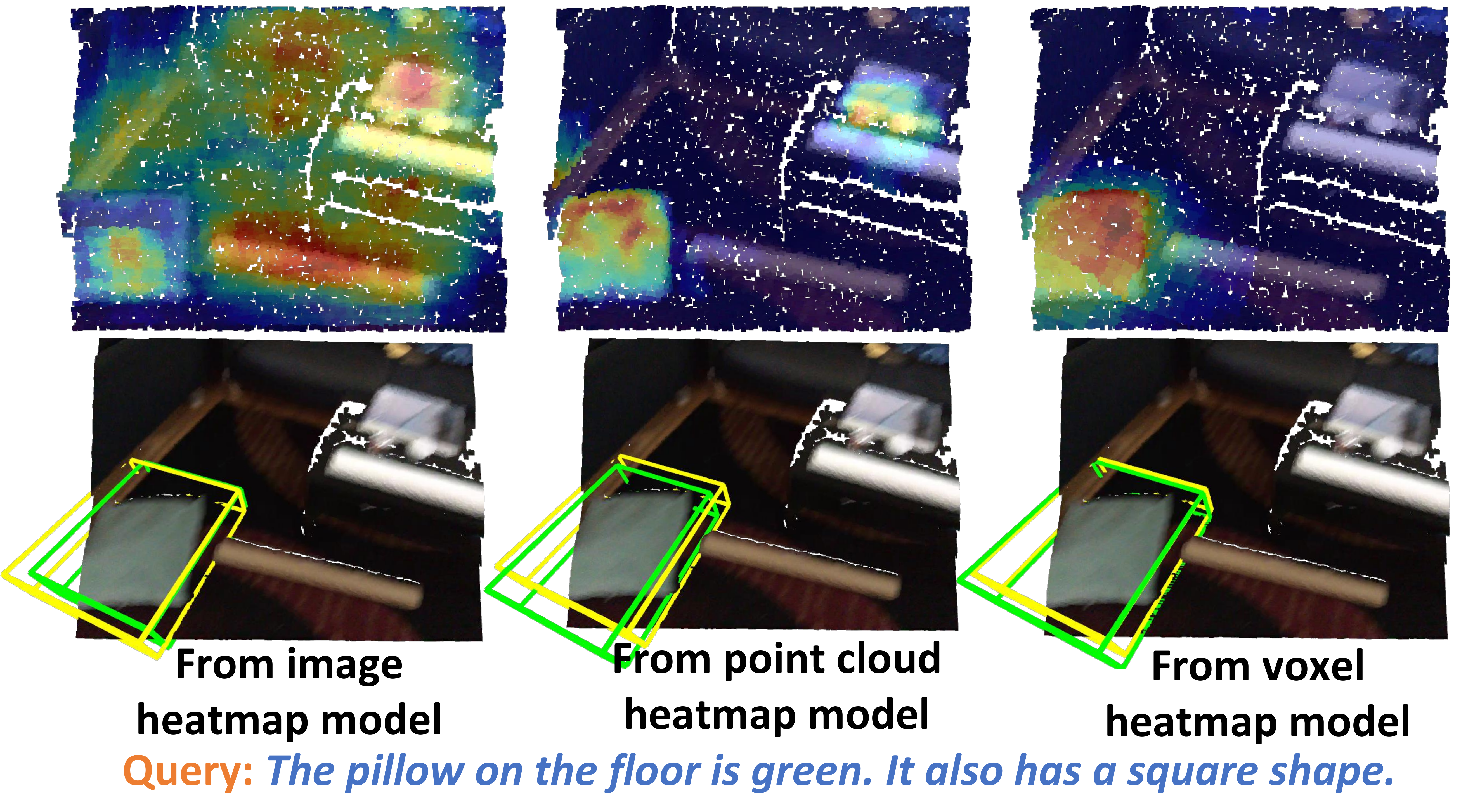}
    \caption{Comparison between different modalities used by the heatmap generation model. Successful predictions are in green; failures in red; and ground-truth in yellow.}
    \label{diff_modality}
\end{figure}

We show the performances of the three models using different modalities in the second row block of Table~\ref{table_abalation}. Comparison shows our method using the voxel-based heatmap achieves the best result among the three optional configurations. 
In terms of efficiency, the voxel-based model can run at 10.0 fps, slightly lower than the frame-rate of the image-based model (12.5 fps) but much higher than that of the point cloud model (5.3 fps). Qualitative examples are given in Figure~\ref{diff_modality}. Our voxel-based model can learn a more concentrated heatmap than models using the other two modalities. 
With such a heatmap, the network can extract context-aware features from the regions relevant to the target object, and hence precisely infer the bounding box of the target object.

\textbf{Without proposal refinement:}
The grounding performance can be increased by a moderate margin on both datasets with the additional module designed for refining the final proposals. 
Comparison against results without this refinement module is shown in the bottom block in Table~\ref{table_abalation}.

\section{Conclusions}
This paper presents a novel task of 3D visual grounding in single-view RGBD images. 
A bottom-up method is proposed that matches the query sentence to visual features at both voxel level and object level. 
We also contribute a large-scale referring expression dataset, SUNRefer, on more than 7,000 single-view RGBD images corresponding to $38,495$ descriptions in total.
Extensive experiments on both the SUNRefer and ScanRefer datasets show that the proposed method significantly outperforms previous methods. 


\section{Acknowledgements}
This work was partially supported by National Key Research and Development Program of China (No.2020YFC2003902). It was supported in part by the Key Area R\&D Program of Guangdong Province with grant No. 2018B030338001, by the National Key R\&D Program of China with grant No. 2018YFB1800800, by Shenzhen Outstanding Talents Training Fund, and by Guangdong Research Project No. 2017ZT07X152, the National Natural Science Foundation of China 61902334, Shenzhen Fundamental Research (General Project) JCYJ20190814112007258.

{\small
\bibliographystyle{ieee_fullname}
\bibliography{egbib}
}

\end{document}